# Automatic Clustering with Single Optimal Solution

Karteeka Pavan K (Corresponding author)

Rayapati Venkata Ranga Rao and Jagarlamudi Chadramouli College of Engineering

Guntur, India – 522 019E-mail: karteeka@yahoo.com

Allam Appa Rao

Jawaharlal Nehru Technological University, Kakinada, India
E-mail: apparaoallam@gmail.com

A.V. Dattatreya Rao

Department of Statistics, Acharya Nagarjuna University, Guntur, India
E-mail: avdrao@gmail.com

**Abstract**

Determining optimal number of clusters in a dataset is a challenging task. Though some methods are available, there is no algorithm that produces unique clustering solution. The paper proposes an Automatic Merging for Single Optimal Solution (AMSOS) which aims to generate unique and nearly optimal clusters for the given datasets automatically. The AMSOS is iteratively merges the closest clusters automatically by validating with cluster validity measure to find single and nearly optimal clusters for the given data set. Experiments on both synthetic and real data have proved that the proposed algorithm finds single and nearly optimal clustering structure in terms of number of clusters, compactness and separation.

## 1. Introduction

Jain (2010) specified that the partitional clustering technique, k-means, is the most computationally simple and efficient clustering method. Wu et al. (2008) have shown that the k-means was one of the top ten algorithms in data mining. Although the k-means method has a number of advantages over other data clustering methods, the specification of number of clusters is a priori, which is usually unknown.

Discovering an optimal number of clusters in a large data set is usually a challenging task. Jain (2010) shown a number of methods to determine k in k-means type algorithms.. Cheung (2005) studied a rival penalized competitive learning algorithm, and Xu ( 1997, 1996) has demonstrated a very good result in finding the cluster number. The algorithm is formulated by learning the parameters of a mixture model through the maximization of a weighted likelihood function. In the learning process, some initial seed centers move to the genuine positions of the cluster centers in a data set, and other redundant seed points will stay at the boundaries or outside of the clusters. Guo et.al (2002) have provided a unified algorithm for both unsupervised and supervised learning for solving the problem of selection of the cluster number. Lee and Antonsson (2000) used an evolutionary method to dynamically cluster a data set. Sarkar,et al,. (1997) and Fogel, Owens, and Walsh (1966) are proposed an approach to dynamically cluster a data set using evolutionary programming, where two fitness functions are simultaneously optimized: one gives the optimal number of clusters, whereas the other leads to a proper identification of each cluster's centroid. Recently Swagatam Das and Ajith Abraham (2008) proposed an Automatic Clustering using Differential

Evolution (ACDE) algorithm by introducing a new chromosome representation. The majority of these methods to determine the best number of clusters may not work very well in practice. The clustering algorithms are require to be run several times for good solution, and model-based methods, such as cross-validation and penalized likelihood estimation, are computationally expensive. All these algorithms cause different solutions in different independent runs. There is no single clustering algorithm that finds unique set of optimal clusters automatically. This paper proposes a new clustering algorithm Automatic Merging for Single Optimal Solution (AMSOS) is a parameter free, simple finds unique and nearly optimal clusters for large data set automatically.

The proposed Automatic Merging for Single Optimal Solution (AMSOS) is a two-phase iterative procedure. Karteeka et al. (2010) have proposed Single Pass Seed Selection algorithm to initialize seeds for k-means. In the first phase, it produces single set of optimal clusters by initializing initial seeds using SPSS . In the second phase, iteratively a low probability cluster is merged with its closest cluster using average linkage after validating with cluster validity metric. The proposed AMSOS is aimed to meet requirements as 1) nearly optimal clusters and 2) unique clustering solution. Experiments on both synthetic and real data sets from UCI prove that the proposed algorithm finds nearly optimal results in terms of compactness and separation.

Section (2) deals with formulation of the proposed algorithm, while section (3) illustrates the effectiveness of the new algorithm experimenting results on synthetic, real, and micro array data sets. Comments on the results of AMSOS are included in Section (4) and finally concluding remarks are included in section (5).

**2 Materials and Methods**

Let P = {P1, P2,… , Pm} be a set of m objects in which each object Pi is represented as[pi,1,pi,2,…pi,n] where n is the number of features .

*2.1 Automatic Merging for Single Optimal Solution*

The choice of the initial seeds was done by SPSS in AMSOS. The SPSS is an optimal seed selection algorithm that produces unique set of initial seeds to k-means type algorithms and is a modification to k-means++ proposed by Arthu and Vassilvitskii (2007). Thus the AMSOS produces single and nearly optimal clustering solution. In the clustering literature Pal and Bezdek (1995) reported that the number of clusters in the data set is in the range from 2 to $\sqrt{m}$ . The AMSOS algorithm first finds kmax $=\sqrt{m}$ clusters using k-means and iteratively merges the lower probability cluster with its closest cluster according to average linkage and validates the merging result using Rand Index which is proposed by Rand (1971). The probability of cluster A is as follows.

$$P(A) = \frac{|A|}{|X|}$$

where $|A|$ is number of elements belongs to cluster A and $|X|$ is the total number of elements in the original dataset, X. The distance between two clusters is measured using average linkage i.e. the distance between two clusters, D( $C_i$,$C_j$) is computed as the average distance between elements from the first cluster and elements from the second cluster and as shown in the following equation.

$$D(C_i, C_j) = \frac{1}{n_i n_j} \sum_{x \in C_i} \sum_{x' \in C_j} |x - x'|$$

Steps:

1. Initialize kmax, number of clusters to the square root of total number of objects
2. Assign kmax objects using SPSS to the cluster centroids
3. Find the clusters using k-means
4. Compute Rand index
5. Find a cluster that has least probability and merge with its closest cluster. Recompute centroids, Rand index and decrement the number of clusters by one. If the newly computed Rand index is greater than the previous Rand index, then update Rand Index, number of clusters and cluster centroids with the newly computed values.
6. If step 5 has been executed for each and every cluster, then go to step7, otherwise got to step5.
7. If there is no change in number of clusters, then stop, otherwise go to step2

*2.2 Data sets*

The efficiency of new algorithms are evaluated by conducting experiments on four artificial data sets, three real datasets down loaded from the web site UCI and two microarray data sets (two yeast data sets) downloaded from http://www.cs. washington.edu/homes/kayee/cluster (Yeung 2001).

The real data sets used:

1. Iris plants database (n = 150, d = 4, K = 3)
2. Glass (n = 214, d = 9, K = 6)
3. Wine (n = 178, d = 13, K = 3)

The real microarray data sets used:

1. The yeast cell cycle data (Cho et al., 1998) showed the fluctuation of expression levels of approximately 6000 genes over two cell cycles (17 time points). We used two different subsets of this data with independent external criteria. The first subset (the 5-phase criterion) consists of 384 genes whose expression levels peak at different time points corresponding to the five phases of cell cycle (Cho et al., 1998). We expect clustering results to approximate this five class partition. Hence, we used the 384 genes with the 5- phase criterion as one of our data sets.
2. The second subset (the MIPS criterion) consists of 237 genes corresponding to four categories in the MIPS database (Mewes et al., 1999). The four categories (DNA synthesis and replication, organization of centrosome, nitrogen and sulphur metabolism, and ribosomal proteins) were shown to be reflected in clusters from the yeast cell cycle data (Tavazoie et al., 1999).

The four synthetic data sets from $N_p(\mu, \sum)$ with specified mean vector and variance covariance matrix are as follows.

1. Number of elements, m=350, number of attributes, n=3, number of clusters, k =2 with

$$\mu_1 = \begin{pmatrix} 2 \\ 3 \\ 4 \end{pmatrix} \quad \Sigma_1 = \begin{pmatrix} 1 & 0.5 & 0.3333 \\ & 1 & 0.6667 \\ & & 1 \end{pmatrix} \quad \mu_2 = \begin{pmatrix} 7 \\ 6 \\ 9 \end{pmatrix} \quad \Sigma_2 = \begin{pmatrix} 1 & 1 & 1 \\ & 2 & 2 \\ & & 3 \end{pmatrix}$$

2. The data set with m=400, n=3, clusters=4 with

$$\mu1 = \begin{pmatrix} -1 \\ -1 \end{pmatrix} \Sigma1 = \begin{pmatrix} 0.65 & 0 \\ 0 & 0.65 \end{pmatrix} \mu2 = \begin{pmatrix} 2 \\ 2 \end{pmatrix} \Sigma2 = \begin{pmatrix} 1 & 0.7 \\ 0 & 1 \end{pmatrix} \mu3 = \begin{pmatrix} -3 \\ +3 \end{pmatrix} \Sigma3 = \begin{pmatrix} 0.78 & 0 \\ 0 & 0.78 \end{pmatrix} \mu4 = \begin{pmatrix} -6 \\ +4 \end{pmatrix} \Sigma4 = \begin{pmatrix} 0.5 & 0 \\ 0 & 0.5 \end{pmatrix}$$

3. m=300, n=2, k=3;

$$\mu1 = \begin{pmatrix} -1 \\ -1 \end{pmatrix} \Sigma1 = \begin{pmatrix} 1 & 0 \\ 0 & 1 \end{pmatrix} \mu2 = \begin{pmatrix} 2 \\ 2 \end{pmatrix} \Sigma2 = \begin{pmatrix} 1 & 0 \\ 0 & 1 \end{pmatrix} \mu3 = \begin{pmatrix} -3 \\ +3 \end{pmatrix} \Sigma3 = \begin{pmatrix} 0.7 & 0 \\ 0 & 0.7 \end{pmatrix}$$

4. m=800, n=2, k=6;

$$\mu1 = \begin{pmatrix} -1 \\ -1 \end{pmatrix} \Sigma1 = \begin{pmatrix} 0.65 & 0 \\ 0 & 0.65 \end{pmatrix} \mu2 = \begin{pmatrix} -8 \\ -6 \end{pmatrix} \Sigma2 = \begin{pmatrix} 1 & 0.7 \\ 0 & 1 \end{pmatrix} \mu3 = \begin{pmatrix} -3 \\ +6 \end{pmatrix} \Sigma3 = \begin{pmatrix} 0.2 & 0 \\ 0 & 0.2 \end{pmatrix} \mu4 = \begin{pmatrix} -8 \\ +14 \end{pmatrix} \Sigma4 = \begin{pmatrix} 0.5 & 0 \\ 0 & 0.5 \end{pmatrix}$$

$$\mu5 = \begin{pmatrix} 10 \\ 12 \end{pmatrix} \Sigma5 = \begin{pmatrix} 0.3 & 0 \\ 0 & 0.3 \end{pmatrix} \mu6 = \begin{pmatrix} +14 \\ -14 \end{pmatrix} \Sigma6 = \begin{pmatrix} 0.1 & 0 \\ 0 & 0.1 \end{pmatrix}$$

**3. Experimental Results**

The clustering results of AMSOS are compared with the results of K-means, k-means++, SPSS, a method to obtain robust optimal centroids in a single pass , which is developed by Karteeka (2011), fuzzy-kmeans, and Automatic Clustering using Differential Evolution (ACDE) to determine optimal clusters.

The k-means, k-means++, Fuzzy-k and SPSS algorithms are implemented with the number of clusters as equal to the number of classes in the ground truth.

*3.2 Presentation of Results*

While comparing the performance of AMSOS with the other techniques (k-means, k-means++, fuzzy-k, SPSS, ACDE) we are concentrating on two major issues: 1) quality of the solution as determined by Error rate and with cluster validity measures Rand, has proposed by Rand (1971), Adjusted Rand, DB, proposed by Davis and Bouldin (1979), CS, is proposed by Chou (2004) and Silhouette is proposed by Rousseeuw (1987) 2) ability to find the optimal number of clusters.  Forty independent runs of each algorithm is taken for the algorithms those produce different results in different individual runs. The Rand, Adjusted Rand, DB, CS and Silhouette metrics values and the overall error rate of the mean-of-run solutions provided by the algorithms over the 10 datasets have been provided in Table 1.

The error rate is defined as

$$err = \frac{N_{mis}}{m} *100$$

where $N_{mis}$ is the number of misclassifications and m is the number of elements of data set original data set X. The best performance values and least performance values that found in 40 independent runs of each algorithm on each dataset is tabulated in Table2 and in Table3. Table 4 contains the obtained centroids from AMSOS for the synthetic datasets and their original cluster centroids.

**4. Comments on the results of AMSOS**

- In case of Synthetic1, Synthetic2, Synthetic3, Synthetic4 the AMSOS resulting single, robust, optimal clustering solution with least error rates compared to other existing algorithms.
- Table4 demonstrates that the efficiency of AMSOS in determining optimal centroids, which are very close to the original centroids.

- The Automatic Merging for Single Optimal Solution (AMSOS) performs well on different data sets. The misclassification rate of single clustering solution of the AMSOS is either as same as the minimum error rate found in 40 independent runs or nearly equal with the means of different solutions of the existing algorithms for all the data sets. Table 5 shows the error rates of unique solutions of SPSS, AMSOS and means of error rates of forty independent runs of ACDE, k-means, k-means++, fuzzy-k.
- The proposed AMSOS shows an improved performance of 30% over ACDE in terms of error rate.
- For the micro array data sets it is resulting single solution with error rates 35.44, 44.01 for the yeast1 and yeast2.
- The poor performance of AMSOS can be seen in the data sets which contains non separable overlapped clusters.
- The AMSOS produces 82.21% qualitative clusters in terms of Rand validity measure.
- The quality of AMSOS on all data sets is 72.15% in terms of silhouette measure.

Note: Results of CS, ARI, etc., are very much in agreement with above all observations in the performance of AMSOS, hence detailed note with respect to them is not provided to avoid duplication.

## 5. Conclusions

AMSOS is totally a non-parameter procedure. Unlike the most of the algorithms, it does not require any heuristic parameter values in advance, though it requires k as input the output does not depends on the input. Being the high density point is the first seed, the SPSS (Karteeka 2011) avoids different results that occur from random selection of initial seeds. For the remaining seeds it follows k-means++. The algorithm is insensitive to outliers in seed selection. Thus the AMSOS in combination with SPSS centroids is outlier insensitive and results in single clustering solution. The table 4 demonstrated that the proposed algorithm produce clustering result with optimal centroids and with single solution.

**References**

Arthu, D. and Vassilvitskii, S. (2007) , "K-means++: The advantages of careful seeding, proceeding of the 18th Annual ACM-SIAM Symposium of Discrete Analysis",7-9, ACM  Press, New Orleans, Louisiana, pp: 1027-1035.

http://portal.acm.org/citation.cfm?id=1283494, www.stanford.edu/~darthur/kMeansPlusPlus.pdf

Cheung, Y. (2005) , "Maximum Weighted Likelihood via Rival Penalized EM for Density Mixture Clustering with Automatic Model Selection," IEEE Trans. Knowledge and Data Eng., vol. 17, no. 6, pp. 750-761.

DOI: 10.1109/TKDE.2005.97 , http://ieeexplore.ieee.org/xpl/freeabs_all.jsp?arnumber=1423976

Cho, R.J., Campbell, M.J., Winzeler, E.A., Steinmetz, L., Conway, A., Wodicka, L., Wolfsberg, T.G., Gabrielian, A.E., Landsman, D., Lockhart, D.J., and Davis, R.W. (1998)  , "A genome-wide transcriptional analysis of the mitotic cell cycle," Mol. Cell, vol. 2, no. 1, pp. 65–73.

http://computableplant.ics.uci.edu/ICS277C/papers/pathways/Cell%20cycle/Cho_cellcycle_data98.pdf

Chou, C.H., Su, M.C., Lai, E. (2004), "A new cluster validity measure and its application to image compression," Pattern Anal. Appl., vol. 7, no. 2, pp. 205–220

DOI:10.1007/s10044-004-0218-1,

http://www.iis.sinica.edu.tw/~ister/publications/Jounral%20paper/CS%20measure.pdf

Davies, D.L. and Bouldin, D.W. (1979). A cluster separation measure. IEEE Transactions on Pattern Analysis and Machine Intelligence 1979, vol.1, pp.224–227. DOI: 10.1109/TPAMI.1979.4766909 , ieexplore.ieee.org/iel5/34/4766893/04766909.pdf?arnumber=4766909

Fogel, L.J, Owens, A. J. and Walsh, M. J (1996) , "Artificial Intelligence Through Simulated Evolution". New York: Wiley.

Guo, P. Chen, C.L. and Lyu, M.R (2002)"Cluster Number Selection for a Small Set of Samples Using the Bayesian Ying-Yang Model," IEEE Trans. Neural Networks, vol. 13, no. 3, pp. 757-763 DOI: 10.1109/TNN.2002.1000144, ieeexplore.ieee.org/iel5/72/21590/01000144.pdf?arnumber=1000144

Jain, A.K. (2010) "Data Clustering: 50 Years Beyond K-Means" , Pattern Recognition letters, 31, pp 651-666. doi:10.1016/j.patrec.2009.09.011 ,
http://www.cs.ucf.edu/courses/cap6412/fall2009/papers/JainDataClusteringPRL09.pdf

KarteekaPavan K. (2011) Ph.D. thesis titled "Contributions to automatic clustering techniques for optimal structures in microarray data" submitted to Acharya Nagarjuna University , India in 2011.

KarteekaPavan, K. Allam AppaRao, DattatreyaRao, A.V., Sridhar, G.R. (2010). "Single Pass Seed Selection Algorithm for k-means," J.Computer Science 6(1):60-66.
**DOI:** 10.3844/jcssp.2010.60.66 http://www.scipub.org/fulltext/jcs/jcs6160-66.pdf

Lee, C.Y. and Antonsson, E.K. (2000), "Self-adapting vertices for mask-layout synthesis," in Proc. Model. Simul. Microsyst. Conf., M. Laudon and B. Romanowicz, Eds., San Diego, CA, Mar. pp. 83–86.
http://www.design.caltech.edu/Research/Publications/99h.pdf

Mewes, H.W. , Heumann, K., Kaps, A. , Mayer, K. , Pfeiffer, F.stocker, S., and Frishman,D (1999) . MIPS: a database for protein sequience and complete genomes. Nucleic Acids Research, 27:44-48.
http://nar.oxfordjournals.org/content/30/1/31.full.pdf+html

Pal, N.R. and Bezdek, J.C. (1995) "On Cluster Validity for the Fuzzy C-Means Model," IEEE Trans. Fuzzy Systems, vol. 3, no. 3, pp. 370- 379.
DOI: 10.1109/91.413225, ieeexplore.ieee.org/iel4/91/9211/00413225.pdf?arnumber=413225

Rand, W. M. (1971) "Objective criteria for the evaluation of clustering methods. Journal of the American Statistical Association," vol.66, pp.846-850.
DOI: 10.2307/2284239, http://www.jstor.org/pss/2284239

Rousseeuw, P. J. (1987) , "Silhouettes: a graphical aid to the interpretation and validation of cluster analysis". Journal of Computational and Applied Mathematics, vol.20, pp.53–65.
DOI: 10.1016/0377-0427(87)90125-7, portal.acm.org/citation.cfm?id=38772

Sarkar, M., Yegnanarayana, B. and Khemani, D. (1997) , "A clustering algorithm using an evolutionary programming-based approach," Pattern Recognit. Lett., vol. 18, no. 10, pp. 975–986.
DOI: 10.1016/S0167-8655(97)00122-0, http://speech.iiit.ac.in/svlpubs/article/Sarkar1997975.pdf

Swagatam Das, Ajith Abraham (2008) "Automatic Clustering Using An Improved Differential Evolution Algorithm", Ieee Transactions On Systems, Man, And Cybernetics—Part A: Systems And Humans, Vol. 38, No. 1, Pp218-237.
DOI: 10.1109/TSMCA.2007.909595 , www.softcomputing.net/smca-paper1.pdf

Tavazoie, S., Huges, J.D., Campbell, M.J., Cho, R.J. and Church, G.M. (1999) "Systematic determination of genetic network architecture". Nature Genetics, vol.22, pp.281–285.

DOI:10.1038/10343, www.ics.uci.edu/~xhx/courses/CS284A/.../CS284A_Lecture5_cellcycle.pdf

Wu, X., V. Kumar, J.R. Quinlan, J. Ghosh, D.J. Hand and D. Steinberg et al.(2008),Top 10 algorithms in data mining". Knowl. Inform. Syst. J., 14: 1-37.

DOI: 10.1007/s10115-007-0114-2, www.cs.uvm.edu/~icdm/algorithms/10Algorithms-08.pdf -

Xu, L. (1996). "How Many Clusters: A Ying-Yang Machine Based Theory for a Classical Open Problem in Pattern Recognition," Proc. IEEE Int'l Conf. Neural Networks ICNN '96, vol. 3, pp. 1546-1551

DOI:10.1109/ICNN.1996.549130,ieeexplore.ieee.org/iel3/3927/11368/00549130.pdf?arnumber=549130

Xu, L. (1997) "Rival Penalized Competitive Learning, Finite Mixture, and Multisets Clustering," Pattern Recognition Letters, vol. 18, nos. 11- 13, pp. 1167-1178.

DOI:10.1109/IJCNN.1998.687259 ,www.cse.cuhk.edu.hk/~lxu/papers/conf-chapters/XURPCLijcnn98.pdf

Yeung, K.Y. (2001), "Cluster analysis of gene expression data. In PhD thesis University of Washington".
http://faculty.washington.edu/kayee/thesis_kayee.pdf

Table 1. Validity measures with error rates

| Dataset | Algorithm | No. of clusters, k | | Cluster Validity Measures | | | | | | Mean Error rate |
|---|---|---|---|---|---|---|---|---|---|---|
| | | i/p k | o/p k | ARI | RI | HI | SIL | DB | CS | |
| Synthetic1 | k-means | 2 | 2 | 0.92 | 0.96 | 0.92 | 0.839 | 0.467 | 0.645 | 0.236 |
| | k-means++ | | | 0.925 | 0.962 | 0.925 | 0.839 | 0.466 | 0.567 | 1.914 |
| | fuzk | | | 0.899 | 0.95 | 0.9 | 0.839 | 0.468 | 0.52 | 2.571 |
| | SPSS | | | 0.932 | 0.966 | 0.932 | 0.839 | 0.465 | 0.725 | 1.714 |
| | ACDE | 19 | 3.05 | 0.85 | 0.925 | 0.849 | 0.643 | 0.772 | 1.348 | 51.56 |
| | AMSOS | 19 | 2 | 0.932 | 0.966 | 0.932 | 0.839 | 0.465 | 0.75 | 1.714 |
| Synthetic2 | k-means | 4 | 4 | 0.821 | 0.927 | 0.854 | 0.718 | 0.58 | 1.178 | 19.1 |
| | k-means++ | | | 0.883 | 0.953 | 0.907 | 0.776 | 0.519 | 1.21 | 7.16 |
| | fuzk | | | 0.944 | 0.979 | 0.957 | 0.791 | 0.484 | 0.931 | 2.2 |
| | SPSS | | | 0.939 | 0.977 | 0.953 | 0.792 | 0.527 | 0.812 | 2.4 |
| | ACDE | 22 | 5.35 | 0.885 | 0.957 | 0.914 | 0.68 | 0.674 | 1.321 | 58.89 |
| | AMSOS | 22 | 4 | 0.939 | 0.977 | 0.953 | 0.792 | 0.527 | 0.943 | 2.4 |
| Synthetic3 | k-means | 3 | 3 | 0.957 | 0.98 | 0.96 | 0.813 | 0.509 | 0.87 | 2.242 |
| | k-means++ | | | 0.97 | 0.987 | 0.974 | 0.823 | 0.761 | 0.92 | 1 |
| | fuzk | | | 0.97 | 0.987 | 0.974 | 0.823 | 0.5 | 0.96 | 1 |
| | SPSS | | | 0.97 | 0.987 | 0.974 | 0.823 | 0.507 | 0.657 | 1 |
| | ACDE | 17 | 4 | 0.472 | 0.777 | 0.553 | 0.754 | 0.461 | | 83.59 |
| | AMSOS | 17 | 3 | 0.97 | 0.987 | 0.974 | 0.823 | 0.507 | 0.768 | 1 |
| Synthetic4 | k-means | 6 | 6 | 0.816 | 0.941 | 0.882 | 0.82 | 0.407 | 0.72 | 51.27 |

| Dataset | Algorithm | | | | | | | | | |
|---|---|---|---|---|---|---|---|---|---|---|
| | k-means++ | | | 0.958 | 0.988 | 0.976 | 0.932 | 0.222 | 0.62 | 10.96 |
| | fuzk | | | 0.98 | 0.994 | 0.988 | 0.953 | 0.183 | 0.45 | 8.738 |
| | SPSS | | | 1 | 1 | 1 | 0.975 | 0.144 | 0.723 | 0 |
| | ACDE | 28 | 7.9 | 0.979 | 0.994 | 0.989 | 0.878 | 0.308 | 0.359 | 53.21 |
| | AMSOS | 28 | 6 | 1 | 1 | 1 | 0.975 | 0.144 | 0.233 | 0 |
| Iris | k-means | 3 | 3 | 0.774 | 0.892 | 0.785 | 0.804 | 0.463 | 0.607 | 15.77 |
| | k-means++ | | | 0.796 | 0.904 | 0.807 | 0.804 | 0.461 | 0.712 | 13.37 |
| | fuzk | | | 0.788 | 0.899 | 0.798 | 0.803 | 0.46 | 0.658 | 15.33 |
| | SPSS | | | 0.44 | 0.72 | 0.441 | 0.799 | 0.582 | 1.962 | 50.67 |
| | ACDE | 12 | 3.15 | 0.887 | 0.95 | 0.901 | 0.784 | 0.435 | 0.706 | 10.17 |
| | AMSOS | 12 | 2 | 0.568 | 0.776 | 0.553 | 0.952 | 0.233 | 0.402 | 33.33 |
| Wine | k-means | 3 | 3 | 0.295 | 0.675 | 0.35 | 0.694 | 0.569 | 0.612 | 34.58 |
| | k-means++ | | | 0.305 | 0.681 | 0.362 | 0.694 | 0.562 | 0.678 | 33.54 |
| | fuzk | | | 0.34 | 0.7 | 0.401 | 0.696 | 0.566 | 0.753 | 30.34 |
| | SPSS | | | 0.337 | 0.699 | 0.398 | 0.696 | 0.601 | 0.813 | 30.34 |
| | ACDE | 13 | 4.45 | 0.367 | 0.723 | 0.447 | 0.373 | 0.555 | 1.626 | 52.89 |
| | AMSOS | 13 | 2 | 0.197 | 0.593 | 0.186 | 0.714 | 0.644 | 1.024 | 41.01 |
| Glass | k-means | 6 | 6 | 0.245 | 0.691 | 0.382 | 0.507 | 0.901 | 0.967 | 55.86 |
| | k-means++ | | | 0.259 | 0.683 | 0.365 | 0.548 | 0.871 | 1.523 | 56.1 |
| | fuzk | | | 0.241 | 0.72 | 0.44 | 0.293 | 0.998 | 1.613 | 62.29 |
| | SPSS | | | 0.252 | 0.722 | 0.444 | 0.382 | 1.061 | 1.512 | 45.79 |
| | ACDE | 15 | 5.5 | 0.309 | 0.712 | 0.425 | 0.338 | 1.146 | 2.868 | 54.35 |
| | AMSOS | 15 | 5 | 0.27 | 0.669 | 0.337 | 0.639 | 0.95 | 1.366 | 70.56 |
| Yeast1 | k-means | 4 | 4 | 0.497 | 0.765 | 0.53 | 0.466 | 1.5 | 1.439 | 35.74 |
| | k-means++ | | | 0.465 | 0.751 | 0.503 | 0.425 | 1.528 | 1.678 | 37.49 |
| | fuzk | | | 0.43 | 0.734 | 0.468 | 0.37 | 2.012 | 1.679 | 39.18 |
| | SPSS | | | 0.508 | 0.769 | 0.538 | 0.464 | 1.471 | 1.217 | 35.44 |
| | ACDE | 15 | 5.55 | 0.594 | 0.806 | 0.612 | 0.348 | 2.314 | 2.669 | 81.86 |
| | AMSOS | 15 | 4 | 0.508 | 0.769 | 0.538 | 0.464 | 1.471 | 1.515 | 35.44 |
| Yeast2 | k-means | 5 | 5 | 0.447 | 0.803 | 0.607 | 0.438 | 1.307 | 1.721 | 38.35 |
| | k-means++ | | | 0.436 | 0.801 | 0.603 | 0.421 | 1.292 | 1.521 | 40 |
| | fuzk | | | 0.421 | 0.799 | 0.598 | 0.379 | 1.443 | 1.341 | 35.73 |
| | SPSS | | | 0.456 | 0.804 | 0.608 | 0.453 | 1.236 | 2.567 | 43.23 |
| | ACDE | 20 | 6.225 | 0.537 | 0.838 | 0.677 | 0.363 | 1.438 | 2.326 | 44.95 |
| | AMSOS | 20 | 4 | 0.469 | 0.8 | 0.6 | 0.506 | 1.154 | 1.342 | 44.01 |

Table 2. Best Validity indices along with error rate

| Dataset | Algorithm | No. of | Cluster Validity Measures | Minimu |
|---|---|---|---|---|

156

|            |           | clusters, k | | | | | | | | m Error |
|            |           | i/p k | o/p k | ARI | RI | HI | SIL | DB | CS | rate |
|---|---|---|---|---|---|---|---|---|---|---|
| Synthetic1 | k-means   | 2 |   | 0.932 | 0.966 | 0.932 | 0.839 | 0.465 | 0.75 | 1.714 |
|            | k-means++ |   |   | 0.932 | 0.966 | 0.932 | 0.839 | 0.465 | 0.75 | 1.714 |
|            | fuzk      |   |   | 0.899 | 0.95  | 0.9   | 0.839 | 0.468 | 0.732 | 2.571 |
|            | SPSS      |   |   | 0.932 | 0.966 | 0.034 | 0.839 | 0.465 | 0.725 | 1.714 |
|            | ACDE      | 19 | 2 | 1 | 1 | 1 | 0.839 | 0.463 | 2.13 | 0 |
|            | AMSOS     | 19 | 2 | 0.932 | 0.966 | 0.932 | 0.839 | 0.465 | 0.75 | 1.714 |
| Synthetic2 | k-means   | 4 |   | 0.939 | 0.977 | 0.953 | 0.792 | 0.44 | 1.821 | 2.4 |
|            | k-means++ |   |   | 0.939 | 0.977 | 0.953 | 0.792 | 0.44 | 1.805 | 2.4 |
|            | fuzk      |   |   | 0.944 | 0.979 | 0.957 | 0.791 | 0.44 | 0.931 | 2.2 |
|            | SPSS      |   |   | 0.939 | 0.977 | 0.023 | 0.792 | 0.527 | 0.812 | 2.4 |
|            | ACDE      | 22 | 4 | 0.939 | 0.977 | 0.953 | 0.79 | 0.445 | 2.224 | 2.4 |
|            | AMSOS     | 22 | 4 | 0.939 | 0.977 | 0.953 | 0.792 | 0.527 | 0.943 | 2.4 |
| Synthetic3 | k-means   | 3 |   | 0.97 | 0.987 | 0.974 | 0.823 | 0.474 | 0.768 | 1 |
|            | k-means++ |   |   | 0.97 | 0.987 | 0.974 | 0.823 | 0.474 | 1.701 | 1 |
|            | fuzk      |   |   | 0.97 | 0.987 | 0.974 | 0.823 | 0.474 | 0.749 | 1 |
|            | SPSS      |   |   | 0.97 | 0.987 | 0.013 | 0.823 | 0.507 | 0.657 | 1 |
|            | ACDE      | 17 | 4 | 0.566 | 0.836 | 0.671 | 0.872 | 0.19 | 1.764 | 50 |
|            | AMSOS     | 17 | 3 | 0.97 | 0.987 | 0.974 | 0.823 | 0.507 | 0.768 | 1 |
| Synthetic4 | k-means   | 6 |   | 1 | 1 | 1 | 0.975 | 0.139 | 0.759 | 0 |
|            | k-means++ |   |   | 1 | 1 | 1 | 0.975 | 0.142 | 0.403 | 0 |
|            | fuzk      |   |   | 1 | 1 | 1 | 0.975 | 0.127 | 0.412 | 0 |
|            | SPSS      |   |   | 1 | 1 | 0 | 0.975 | 0.144 | 0.723 | 0 |
|            | ACDE      | 28 | 6 | 1 | 1 | 1 | 0.975 | 0.136 | 0.605 | 0 |
|            | AMSOS     | 28 | 6 | 1 | 1 | 1 | 0.975 | 0.144 | 0.233 | 0 |
| Iris       | k-means   | 3 |   | 0.886 | 0.95 | 0.899 | 0.806 | 0.411 | 0.753 | 4 |
|            | k-means++ |   |   | 0.886 | 0.95 | 0.899 | 0.806 | 0.411 | 0.753 | 4 |
|            | fuzk      |   |   | 0.886 | 0.95 | 0.899 | 0.806 | 0.411 | 0.769 | 4 |
|            | SPSS      |   |   | 0.44 | 0.72 | 0.28 | 0.799 | 0.582 | 1.962 | 50.67 |
|            | ACDE      | 12 | 3 | 0.904 | 0.958 | 0.916 | 0.806 | 0.435 | 1.061 | 3.333 |
|            | AMSOS     | 12 | 2 | 0.568 | 0.776 | 0.553 | 0.952 | 0.233 | 0.402 | 33.33 |
| Wine       | k-means   | 3 |   | 0.337 | 0.699 | 0.398 | 0.696 | 0.447 | 0.939 | 30.34 |
|            | k-means++ |   |   | 0.337 | 0.699 | 0.398 | 0.696 | 0.447 | 0.939 | 30.34 |
|            | fuzk      |   |   | 0.347 | 0.704 | 0.408 | 0.696 | 0.488 | 0.929 | 29.78 |
|            | SPSS      |   |   | 0.337 | 0.699 | 0.301 | 0.696 | 0.601 | 0.813 | 30.34 |
|            | ACDE      | 13 | 2 | 0.423 | 0.755 | 0.511 | 0.686 | 0.555 | 3.368 | 28.65 |
|            | AMSOS     | 13 | 2 | 0.197 | 0.593 | 0.186 | 0.714 | 0.644 | 1.024 | 41.01 |

| Glass | k-means | 6 | | 0.287 | 0.728 | 0.456 | 0.656 | 0.744 | 1.917 | 44.86 |
|---|---|---|---|---|---|---|---|---|---|---|
| | k-means++ | | | 0.288 | 0.725 | 0.45 | 0.729 | 0.522 | 1.745 | 46.73 |
| | fuzk | | | 0.263 | 0.733 | 0.467 | 0.317 | 0.883 | 3.995 | 48.13 |
| | SPSS | | | 0.252 | 0.722 | 0.278 | 0.382 | 1.061 | 1.512 | 45.79 |
| | ACDE | 15 | 4 | 0.331 | 0.758 | 0.517 | 0.636 | 1.146 | 3.883 | 37.38 |
| | AMSOS | 15 | 5 | 0.27 | 0.669 | 0.337 | 0.639 | 0.95 | 1.366 | 70.56 |
| Yeast1 | k-means | 4 | | 0.515 | 0.773 | 0.545 | 0.473 | 1.307 | 2.117 | 35.02 |
| | k-means++ | | | 0.515 | 0.772 | 0.545 | 0.473 | 1.376 | 2.006 | 35.02 |
| | fuzk | | | 0.453 | 0.744 | 0.489 | 0.396 | 1.722 | 16.91 | 37.55 |
| | SPSS | | | 0.508 | 0.769 | 0.231 | 0.464 | 1.471 | 2.012 | 35.44 |
| | ACDE | 15 | 3 | 0.661 | 0.838 | 0.675 | 0.418 | 2.314 | 4.311 | 24.47 |
| | AMSOS | 15 | 4 | 0.508 | 0.769 | 0.538 | 0.464 | 1.471 | 1.515 | 35.44 |
| Yeast2 | k-means | 5 | | 0.491 | 0.818 | 0.635 | 0.455 | 1.213 | 1.217 | 27.08 |
| | k-means++ | | | 0.497 | 0.82 | 0.64 | 0.514 | 1.092 | 1.666 | 26.3 |
| | fuzk | | | 0.478 | 0.812 | 0.625 | 0.43 | 1.296 | 6.384 | 27.86 |
| | SPSS | | | 0.456 | 0.804 | 0.196 | 0.453 | 1.236 | 2.567 | 43.23 |
| | ACDE | 20 | 5 | 0.551 | 0.846 | 0.692 | 0.428 | 1.438 | 3.489 | 23.18 |
| | AMSOS | 20 | 4 | 0.469 | 0.8 | 0.6 | 0.506 | 1.154 | 1.342 | 44.01 |

Table3. Least performance values

| Dataset | Algorithm | No. of clusters, k | | Cluster Validity Measures | | | | | | Maximum Error rate |
|---|---|---|---|---|---|---|---|---|---|---|
| | | i/p k | o/p k | ARI | RI | HI | SIL | DB | CS | |
| Synthetic1 | k-means | 2 | | 0.91 | 0.955 | 0.91 | 0.839 | 0.467 | 0.749 | 2.286 |
| | k-means++ | | | 0.91 | 0.955 | 0.91 | 0.839 | 0.467 | 0.749 | 2.286 |
| | fuzk | | | 0.899 | 0.95 | 0.9 | 0.839 | 0.468 | 0.732 | 2.571 |
| | SPSS | | | 0.932 | 0.966 | 0.034 | 0.839 | 0.465 | 0.725 | 1.714 |
| | ACDE | 19 | 4 | 0.699 | 0.849 | 0.697 | 0.44 | 1.275 | 2.13 | 96 |
| | AMSOS | 19 | 2 | 0.932 | 0.966 | 0.932 | 0.839 | 0.465 | 0.75 | 1.714 |
| Synthetic2 | k-means | 4 | | 0.561 | 0.82 | 0.641 | 0.503 | 0.904 | 0.936 | 67 |
| | k-means++ | | | 0.566 | 0.822 | 0.643 | 0.507 | 0.874 | 0.936 | 59.8 |
| | fuzk | | | 0.944 | 0.979 | 0.957 | 0.791 | 0.528 | 0.93 | 2.2 |
| | SPSS | | | 0.939 | 0.977 | 0.023 | 0.792 | 0.527 | 0.812 | 2.4 |
| | ACDE | 22 | 7 | 0.853 | 0.945 | 0.891 | 0.537 | 0.865 | 2.224 | 96.2 |
| | AMSOS | 22 | 4 | 0.939 | 0.977 | 0.953 | 0.792 | 0.527 | 0.943 | 2.4 |
| Synthetic3 | k-means | 3 | | 0.97 | 0.987 | 0.974 | 0.823 | 0.507 | 0.749 | 1 |

|  |  |  |  |  |  |  |  |  |  |
|---|---|---|---|---|---|---|---|---|---|
|  | k-means++ |  |  | 0.432 | 0.718 | 0.436 | 0.44 | 0.962 | 0.749 | 50.67 |
|  | fuzk |  |  | 0.97 | 0.987 | 0.974 | 0.823 | 0.507 | 0.731 | 1 |
|  | SPSS |  |  | 0.97 | 0.987 | 0.013 | 0.823 | 0.507 | 0.657 | 1 |
|  | ACDE | 17 | 4 | -0.049 | 0.382 | -0.235 | 0.203 | 1.078 | 1.764 | 87.5 |
|  | AMSOS | 17 | 3 | 0.97 | 0.987 | 0.974 | 0.823 | 0.507 | 0.768 | 1 |
| Synthetic4 | k-means | 6 |  | 0.574 | 0.851 | 0.703 | 0.504 | 0.727 | 0.233 | 0 |
|  | k-means++ |  |  | 0.832 | 0.951 | 0.902 | 0.791 | 0.523 | 0.233 | 92.63 |
|  | fuzk |  |  | 0.836 | 0.952 | 0.904 | 0.786 | 0.503 | 0.233 | 94.5 |
|  | SPSS |  |  | 1 | 1 | 0 | 0.975 | 0.144 | 0.723 | 0 |
|  | ACDE | 28 | 10 | 0.944 | 0.985 | 0.97 | 0.736 | 0.511 | 0.247 | 93.88 |
|  | AMSOS | 28 | 6 | 1 | 1 | 1 | 0.975 | 0.144 | 0.233 | 0 |
| Iris | k-means | 3 |  | 0.44 | 0.72 | 0.441 | 0.798 | 0.582 | 0.607 | 51.33 |
|  | k-means++ |  |  | 0.44 | 0.72 | 0.441 | 0.798 | 0.582 | 0.607 | 51.33 |
|  | fuzk |  |  | 0.45 | 0.725 | 0.449 | 0.792 | 0.576 | 0.603 | 56 |
|  | SPSS |  |  | 0.44 | 0.72 | 0.28 | 0.799 | 0.582 | 1.962 | 50.67 |
|  | ACDE | 12 | 4 | 0.795 | 0.914 | 0.828 | 0.623 | 0.435 | 0.529 | 62.67 |
|  | AMSOS | 12 | 2 | 0.568 | 0.776 | 0.553 | 0.952 | 0.233 | 0.402 | 33.33 |
| Wine | k-means | 3 |  | 0.217 | 0.628 | 0.256 | 0.692 | 0.608 | 0.78 | 42.7 |
|  | k-means++ |  |  | 0.217 | 0.628 | 0.256 | 0.687 | 0.608 | 0.774 | 42.7 |
|  | fuzk |  |  | 0.332 | 0.696 | 0.392 | 0.695 | 0.601 | 0.914 | 30.9 |
|  | SPSS |  |  | 0.337 | 0.699 | 0.301 | 0.696 | 0.601 | 0.813 | 30.34 |
|  | ACDE | 13 | 8 | 0.338 | 0.668 | 0.336 | 0.053 | 0.555 | 0.647 | 69.66 |
|  | AMSOS | 13 | 2 | 0.197 | 0.593 | 0.186 | 0.714 | 0.644 | 1.024 | 41.01 |
| Glass | k-means | 6 |  | 0.152 | 0.666 | 0.333 | 0.207 | 1.168 | 0.966 | 67.29 |
|  | k-means++ |  |  | 0.189 | 0.626 | 0.252 | 0.356 | 1.023 | 0.722 | 64.95 |
|  | fuzk |  |  | 0.207 | 0.707 | 0.415 | 0.243 | 1.178 | 1.85 | 66.82 |
|  | SPSS |  |  | 0.252 | 0.722 | 0.278 | 0.382 | 1.061 | 1.512 | 45.79 |
|  | ACDE | 15 | 8 | 0.293 | 0.646 | 0.291 | 0.071 | 1.146 | 0.99 | 86.45 |
|  | AMSOS | 15 | 5 | 0.27 | 0.669 | 0.337 | 0.639 | 0.95 | 1.366 | 70.56 |
| Yeast1 | k-means | 4 |  | 0.246 | 0.658 | 0.315 | 0.184 | 1.757 | 1.509 | 80.17 |
|  | k-means++ |  |  | 0.43 | 0.735 | 0.47 | 0.399 | 2.007 | 1.509 | 42.62 |
|  | fuzk |  |  | 0.394 | 0.721 | 0.441 | 0.343 | 2.239 | 6.311 | 80.59 |
|  | SPSS |  |  | 0.508 | 0.769 | 0.231 | 0.464 | 1.471 | 1.217 | 35.44 |
|  | ACDE | 15 | 8 | 0.545 | 0.786 | 0.573 | 0.233 | 2.314 | 0.942 | 97.47 |
|  | AMSOS | 15 | 4 | 0.508 | 0.769 | 0.538 | 0.464 | 1.471 | 1.515 | 35.44 |
| Yeast2 | k-means | 5 |  | 0.361 | 0.784 | 0.568 | 0.339 | 1.489 | 1.53 | 57.03 |
|  | k-means++ |  |  | 0.367 | 0.786 | 0.572 | 0.364 | 1.354 | 1.21 | 57.03 |
|  | fuzk |  |  | 0.369 | 0.769 | 0.538 | 0.319 | 1.819 | 2.201 | 53.65 |
|  | SPSS |  |  | 0.456 | 0.804 | 0.196 | 0.453 | 1.236 | 2.567 | 43.23 |

|      |    |   |       |      |       |       |       |       |       |
|------|----|---|-------|------|-------|-------|-------|-------|-------|
| ACDE | 20 | 8 | 0.513 | 0.83 | 0.659 | 0.238 | 1.438 | 1.713 | 86.46 |
| AMSOS| 20 | 4 | 0.469 | 0.8  | 0.6   | 0.506 | 1.154 | 1.342 | 44.01 |

Table4. AMSOS efficiency in finding optimal Centroids

| Data set | Orignal Centroids | Obtained Centroids by AMSOS |
|---|---|---|
| Synthetic1 | 7   6   9<br>2   3   4 | 7.0783   5.9625   9.0975<br>2.0289   3.0849   4.1540 |
| Synthetic2 | -6   4<br> 2   2<br>-1  -1<br>-3  -3 | -5.9234   4.0052<br> 1.8901   1.9421<br>-0.9611  -1.2146<br>-2.7994   2.9561 |
| Synthetic3 | -3   3<br>-1  -1<br> 2   2 | -3.1959   2.9669<br>-0.8401  -1.1502<br> 2.0215   1.8696 |
| Synthetic4 | -8  14<br>10  12<br>14  -14<br>-1  -1<br>-3   6<br>-8  -6 | -8.0344   14.0421<br>10.0285   12.0065<br>13.9763  -13.9768<br>-1.1876   -0.9205<br>-2.9580    6.0961<br>-7.9533   -6.0640 |

Figure 1: AMSOS efficiency in clustering the Yeast2 dataset

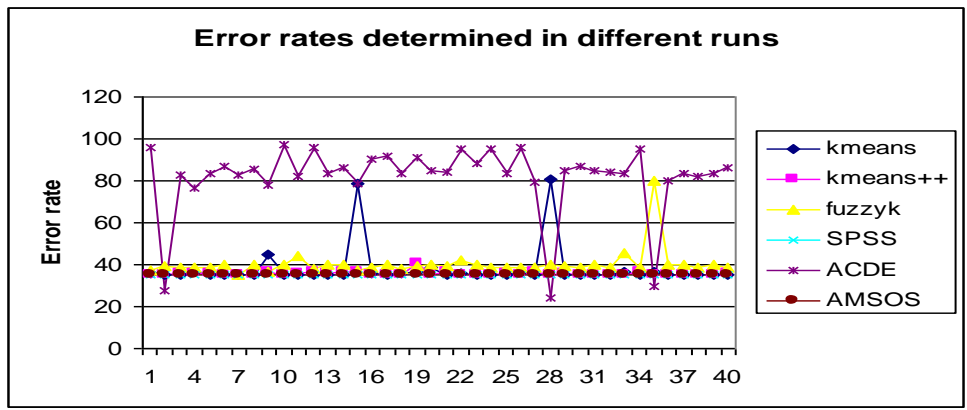

Figure 2: AMSOS efficiency in determining number of clusters for synthetic2 dataset

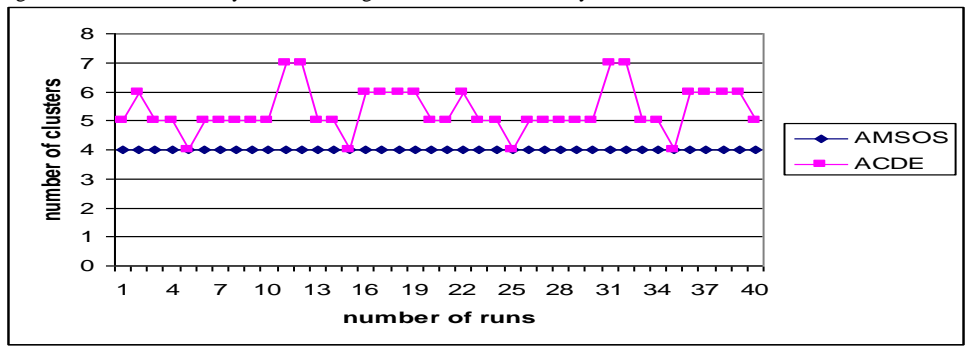

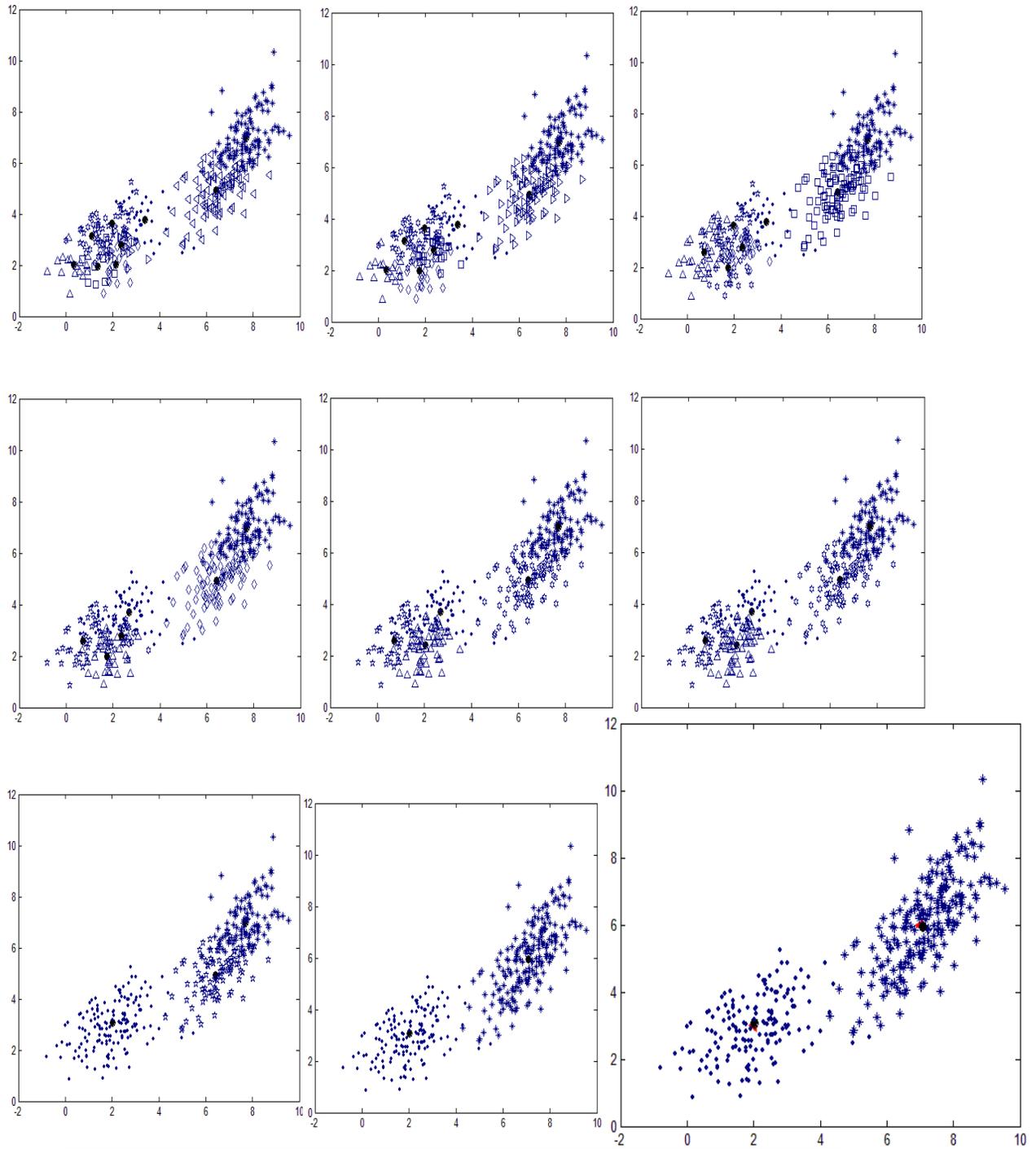

Figure 3. The results obtained by AMSOS for the Synthetic1 data set when initial k=9. Starting with initial clusters to final clusters and their obtained centers. The obtained centers are marked with '●' whereas original centers are marked in red colored triangles.